\documentclass{article}
    
\IfFileExists{neurips_2026.sty}{%
  \usepackage[eandd, final]{neurips_2026}%
}{%
  \usepackage[margin=1in]{geometry}%
  \usepackage[numbers]{natbib}%
}
\usepackage{eso-pic}
\usepackage[utf8]{inputenc}
\usepackage[T1]{fontenc}
\usepackage{hyperref}
\usepackage{url}
\usepackage{booktabs}
\usepackage{amsfonts}
\usepackage{amsmath}
\usepackage{amssymb}
\usepackage{nicefrac}
\usepackage{microtype}
\usepackage{xcolor}
\usepackage{graphicx}
\usepackage{array}
\usepackage{tabularx}
\usepackage{float}

\newcolumntype{Y}{>{\raggedright\arraybackslash}X}
\newcolumntype{L}[1]{>{\raggedright\arraybackslash}p{#1}}
\newcolumntype{R}[1]{>{\raggedleft\arraybackslash}p{#1}}
\newcolumntype{P}{>{\hsize=.85\hsize\raggedright\arraybackslash}X}
\newcolumntype{T}{>{\hsize=1.45\hsize\raggedright\arraybackslash}X}

\title{OTel: Open Telco AI Datasets, Benchmarks, and Models}

\author{
  Farbod Tavakkoli\textsuperscript{1}\thanks{Corresponding author.}\quad
  Gregory Diamos\textsuperscript{2}\quad
  Kenneth Church\textsuperscript{3} \\
  \textbf{Roderic Paulk}\textsuperscript{1}\quad
  \textbf{Ashish Vaswani}\textsuperscript{4}\quad
  \textbf{David Kanter}\textsuperscript{5} \\
  \textbf{Jorden Terrazas}\textsuperscript{1}\quad
  \textbf{Mark Austin}\textsuperscript{1}\quad
  \textbf{Imtiaz Karim}\textsuperscript{6} \\
  \textbf{Syed Ali Raza Zaidi}\textsuperscript{7}\quad
  \textbf{Lina Bariah}\textsuperscript{8} \\
  \textbf{Louis Powell}\textsuperscript{9}\quad
  \textbf{Ali Maatouk}\textsuperscript{10} \\[10pt]
  {\normalfont\small
  \textsuperscript{1}AT\&T Chief Data Office\quad
  \textsuperscript{2}RelationalAI\quad
  \textsuperscript{3}Northeastern University} \\
  {\normalfont\small
  \textsuperscript{4}Essential AI\quad
  \textsuperscript{5}MLCommons\quad
  \textsuperscript{6}The University of Texas at Dallas} \\
  {\normalfont\small
  \textsuperscript{7}University of Leeds\quad
  \textsuperscript{8}Khalifa University\quad
  \textsuperscript{9}GSMA} \\
  {\normalfont\small
  \textsuperscript{10}Yale University}
}

\begin{document}

\AddToShipoutPictureFG*{%
  \AtPageUpperLeft{%
    \raisebox{-1.6cm}{\makebox[\paperwidth]{%
      \small\bfseries Accepted as a Spotlight paper at NeurIPS 2026, Evaluations \& Datasets Track}}}}

\maketitle

\begin{abstract}
We present Open Telco (OTel), an open telecom AI resource that releases derived telecom datasets for retrieval, reranking, instruction tuning, and safety/abstention, together with 30 full-parameter post-trained baselines spanning 10 embedding models, 3 rerankers, and 17 language models. The community has already engaged substantially with the resource: as of May 3, 2026, the released models have been downloaded over 16 million times and the project has received 157+ pieces of media coverage worldwide. Building on prior open telecom datasets and benchmarks, OTel provides documented telecom data sources, held-out evaluation partitions, trained embedding models, rerankers, context-grounded LLMs, and safety/abstention data in one unified resource. Each baseline starts from an open-weight model and is post-trained on OTel-derived data using an open training recipe, then evaluated on held-out OTel evaluation partitions. OTel post-training improves performance across all three model families: embedding retrieval reaches 93.1\% NDCG@10, reranking reaches 0.947 MRR@10, and language-model correctness reaches 87.8\%. We release OTel as a reproducible starting point and invite the community to expand the data, improve embedding and reranking models, and build stronger context-grounded telecom LLMs.
\end{abstract}

\vspace{-10pt}
\section{Introduction}

Open Telco (OTel)\footnote{\url{https://huggingface.co/farbodtavakkoli}}\footnote{\url{https://github.com/farbodtavakkoli/OTel}} is an open telecom AI resource designed to help researchers and practitioners train, evaluate, and improve retrieval, reranking, context-grounded generation, and abstention for telecommunications. The resource combines derived telecom datasets, held-out evaluation partitions, and 30 released full-parameter post-trained baselines across embedding models, rerankers, and language models. Our goal is to provide a reproducible starting point for the community rather than a final answer: the released baselines are intended to make it easier for others to compare methods, identify weaknesses, and improve telecom retrieval, reranking, context-grounded generation, and abstention.

The paper makes five points:
\begin{enumerate}
    \item \textbf{Resource.} OTel is a new open telecom AI resource that builds on prior telecom benchmarks and documents data sources, tasks, and derived dataset formats.
    \item \textbf{Community use.} The resource has already seen broad community engagement, with over 16 million model downloads and 157+ media mentions as of May 3, 2026.
    \item \textbf{Reference baselines.} OTel releases Hugging Face reference baselines and invites the community to improve on them.
    \item \textbf{Headline performance.} The released baselines reach 93.1\% NDCG@10 for retrieval, 0.947 MRR@10 for reranking, and 87.8\% LLM correctness.
    \item \textbf{Fine-tuning and scale.} The results show that OTel fine-tuning improves performance across model families and that larger models generally raise the upper envelope.
\end{enumerate}

There has been considerable recent work on telecom benchmarks and datasets~\citep{karim-etal-2023-spec5g}. TeleQnA evaluates telecom question answering and standards understanding \citep{teleqna}. ORAN-Bench and srsRAN-Bench evaluate O-RAN specifications and open-source 5G code understanding \citep{oranbench}. Additional benchmarks target 3GPP group classification, telecom table reasoning, telecom mathematical reasoning, and 5G root-cause analysis \citep{threegpp,teletables,telemath,telelogs}. The GSMA Open Telco AI Leaderboard~\citep{gsma_telco_leaderboard} brings these efforts into a shared benchmark view for telecom AI.

OTel builds on this prior work by contributing a unified open resource for training and evaluating telecom RAG components across multiple telecom domains and model families. We use \emph{OTel source corpus} to refer to the publicly available telecom documents used as inputs, including 3GPP specifications, GSMA documents, O-RAN documents, RFCs, whitepapers, academic papers, and web-derived telecom material. We use \emph{OTel derived dataset family} to refer to the cleaned and structured examples released as \texttt{OTel-Embedding}, \texttt{OTel-Reranker}, \texttt{OTel-LLM}, and \texttt{OTel-Safety}. We use \emph{OTel evaluation partitions} to refer to held-out splits from these derived datasets, and \emph{OTel model family} to refer to the 30 released full-parameter post-trained baselines.

This paper presents OTel as an Evaluations \& Datasets contribution organized around three assets:
\begin{enumerate}
    \item A derived OTel dataset family built from heterogeneous telecom sources and cleaned from roughly 1.1M raw training points to a higher-quality final corpus, released as \texttt{OTel-Embedding}, \texttt{OTel-Reranker}, \texttt{OTel-LLM}, and \texttt{OTel-Safety}.
    \item A family of 30 released full-parameter post-trained OTel baselines following the RAG pipeline, comprising 10 embedding models, 3 rerankers, and 17 LLMs. The released model naming convention is \texttt{OTel-Embedding-\{model size\}} for embedding models, \texttt{OTel-Reranker-\{model size\}} for rerankers, and \texttt{OTel-LLM-\{model size\}-\{IT or Reasoning\}} for LLMs; auxiliary safety variants use \texttt{OTel-LLM-\{model size\}-Safety} and are discussed in the appendix.
    \item A reproducible evaluation setup for telecom RAG components, using held-out OTel evaluation partitions and model-family-specific metrics, namely LLM-as-judge correctness for LLMs, NDCG@10 for embeddings, and MRR@10 for rerankers.
\end{enumerate}

As of May 3, 2026, the released OTel models have been downloaded over 16 million times and the project has received 157+ pieces of media coverage, providing an early signal of community demand for open telecom AI resources. We release OTel to support continued work on telecom retrieval, reranking, context-grounded generation, and safety, and to invite the community to improve on the baselines reported here.

\section{Related Work and Benchmark Context}

Telecom benchmarking has expanded rapidly during the last two years and now spans several complementary tasks. TeleQnA measures broad telecom knowledge and standards understanding \citep{teleqna}. ORAN-Bench and srsRAN-Bench probe O-RAN and open-source 5G implementation knowledge \citep{oranbench}. 3GPP-TSG tests working-group classification \citep{threegpp}. TeleTables focuses on reasoning over technical tables \citep{teletables}. TeleMath measures domain-specific mathematical reasoning \citep{telemath}. TeleLogs targets 5G root-cause analysis \citep{telelogs}. The GSMA Open Telco AI Leaderboard~\citep{gsma_telco_leaderboard} is an important step because it consolidates these benchmark tasks into a unified evaluation view for telecom AI.

Viewed as part of the broader dataset and post-training ecosystem, OTel connects retrieval data, reranking supervision, instruction-tuning data, safety/abstention data, and released post-trained baselines that are often studied separately.

\begin{table}[!htbp]
\centering
\small
\setlength{\tabcolsep}{6pt}
\renewcommand{\arraystretch}{1.15}
\begin{tabularx}{\textwidth}{@{}L{3.0cm} P P P T@{}}
\toprule
& \multicolumn{3}{c}{Prior Work}
& \multicolumn{1}{c}{This Work} \\
\cmidrule(lr){2-4}\cmidrule(lr){5-5}
\textbf{Data sources \& tasks}
& \href{https://huggingface.co/datasets/AliMaatouk/Tele-Data}{Tele-Data}
& \href{https://huggingface.co/datasets/NetoAISolutions/TEmbed}{TEmbed}
& GSMA Leaderboard
& OTel \\
\midrule
Telecom scope
& Telecom QA
& Telecom data
& Benchmark tasks
& Multi-domain telecom data \\
\addlinespace[2pt]
Retrieval chunks
& \(\times\)
& \(\times\)
& Eval-only
& \(\checkmark\) explicit chunks \\
\addlinespace[2pt]
Reranker labels
& \(\times\)
& \(\times\)
& Eval-only
& \(\checkmark\) relevance labels \\
\addlinespace[2pt]
LLM instruction data
& \(\checkmark\) QA-oriented
& \(\checkmark\) QA-oriented
& Eval-only
& \(\checkmark\) grounded prompts \\
\addlinespace[2pt]
Safety
& \(\times\)
& \(\times\)
& \(\times\)
& \(\checkmark\) abstention data \\
\addlinespace[2pt]
Held-out eval partitions
& Splits
& Splits
& Benchmark entries
& \(\checkmark\) OTel partitions \\
\addlinespace[2pt]
Post-trained baselines
& \(\times\)
& \(\times\)
& \(\times\)
& \(\checkmark\) 30 baselines \\
\bottomrule
\end{tabularx}
\caption{Data sources and tasks covered by prior open telecom resources and OTel. Retrieval chunks indicate explicit positive/negative retrieval chunks. The comparison shows how OTel builds on prior work by covering the retrieval--reranking--generation workflow and safety/abstention data; it is not intended to diminish earlier datasets or leaderboards.}
\label{tab:resource-comparison}
\end{table}

Table~\ref{tab:resource-comparison} summarizes the role of OTel in this ecosystem. Earlier open datasets such as Tele-Data and TEmbed helped make telecom training data more accessible, while the GSMA leaderboard provides a shared benchmark interface for several telecom tasks. OTel is complementary to these efforts: it contributes aligned open training resources and full-parameter post-trained OTel baselines for studying retrieval, reranking, and context-grounded generation systematically. This positioning is similar in spirit to domain evaluation ecosystems in medicine \citep{medqa}, law \citep{legalbench}, and finance \citep{finben}, where shared resources and baselines make it easier for the community to compare methods and improve over time.

\section{The OTel Resource}

Following the terminology introduced above, the OTel source corpus consists of public telecom documents and contributor-provided telecom examples, while the released OTel artifacts are derived, cleaned, and structured training and evaluation examples. The raw public source documents are not themselves our release contribution; rather, our contribution is the curated OTel derived dataset family produced from these sources for retrieval, reranking, context-grounded generation, and safety.

The OTel resource was curated by more than 100 domain experts from industry and academia. The source corpus spans six source categories: GSMA permanent reference documents, 3GPP specifications, O-RAN documents, RFCs, telecom-specific topical material such as eSIM and roaming, and industry whitepapers and telecom academic papers.

\begin{table}[!htbp]
\centering
\small
\setlength{\tabcolsep}{6pt}
\renewcommand{\arraystretch}{1.12}
\begin{tabularx}{\textwidth}{L{2.7cm} Y R{1.8cm}}
\toprule
Contributor & Source domain & Samples \\
\midrule
Yale University & arXiv telecom papers, 3GPP standards, telecom Wikipedia articles, and telecom-related Common Crawl pages & 681,172 \\
NetoAI & RFC series & 100,751 \\
GSMA & PRDs, Discover portal material, and mixed telecom documents & 158,006 \\
Khalifa University & Industry whitepapers & 62,000 \\
University of Leeds & O-RAN specifications across working groups 1--2, 4--10 & 58,565 \\
Purdue University and UT Dallas & O-RAN documents across working groups & 42,000 \\
\midrule
Total & Raw training points before cleaning & \textasciitilde 1,102,494 \\
\bottomrule
\end{tabularx}
\caption{Contributor summary for the raw OTel corpus.}
\label{tab:contributors}
\end{table}

The incoming data had two distinct formats. Yale contributed roughly 680K question-answer-source triples from telecom papers, standards, Wikipedia, and web-derived telecom pages, which OTel converted into retrieval-ready supervision through enrichment and cleaning. These triples did not include positive or negative retrieval passages, and source documents can be hundreds of thousands of tokens. The remaining contributors provided roughly 420K structured examples with explicit question, positive chunks, negative chunks, answer, and source fields.

To convert the Yale data into retrieval-ready supervision, we used a six-stage enrichment pipeline. First, an ETL stage grouped QA pairs by source document and joined them to the full document text. Second, we created candidate passages via sliding windows and semantic chunking. Third, a retrieval pipeline mined hard negatives both within the source document and across other documents, followed by reranker rescoring. Fourth, we used fact-grounded checks over decomposed answer claims rather than only answer similarity. Fifth, we selected the minimal sufficient context through greedy minimization over top-1, top-3, and all-passage contexts. Sixth, we formatted the verified results into training structures suitable for Multiple Negatives Ranking Loss and related retrieval objectives.

After enrichment, the Yale data was merged with the structured submissions from the remaining contributors, producing roughly 1.1M training points before aggressive filtering. We then performed a shard-based data-quality experiment in which separate full-parameter embedding models were trained on independent shards. Performance varied substantially, from 71.1\% Acc@1 on the best shard to 3.9\% Acc@1 on the noisiest shard, confirming that the raw data required significant cleaning.

The final cleaning pipeline applied four filters: heuristic filtering, reranker-based semantic filtering, embedding-based semantic filtering, and deduplication. For the Yale subset, this reduced 680K examples to 220,334. For the remaining contributors, the same process reduced 420K examples to 106,433. The final retained corpus therefore contains 326,767 higher-confidence examples.

\begin{table}[!htbp]
\centering
\small
\begin{tabular}{lrr}
\toprule
Dataset split & Raw size & Final size \\
\midrule
Yale enrichment path & 680,000 & 220,334 \\
Other contributors & 420,000 & 106,433 \\
\midrule
Total & \textasciitilde 1,100,000 & 326,767 \\
\bottomrule
\end{tabular}
\caption{Attrition through the OTel cleaning pipeline.}
\label{tab:attrition}
\end{table}

The retained corpus is released in four model-family-specific formats ordered by the RAG workflow: \texttt{OTel-Embedding}, \texttt{OTel-Reranker}, \texttt{OTel-LLM}, and \texttt{OTel-Safety}. The first three support the 30 released baseline models; \texttt{OTel-Safety} is released separately and supports the appendix-only abstention variants. Because the sources are heterogeneous, release documentation describes provenance, redistribution assumptions, and source-specific constraints at the asset level.

\begin{table}[!htbp]
\centering
\small
\setlength{\tabcolsep}{5pt}
\renewcommand{\arraystretch}{1.22}
\begin{tabularx}{\textwidth}{L{1.9cm} L{3.0cm} Y L{3.4cm}}
\toprule
RAG role & Dataset on Hugging Face & Task & Key fields \\
\midrule
Retrieval
& \texttt{OTel-Embedding}
& Retrieve relevant telecom passages from an anchor/query.
& \texttt{anchor}, \texttt{positive}, \texttt{negative\_1}--\texttt{negative\_5} \\
Reranking
& \texttt{OTel-Reranker}
& Score query-passage relevance for cross-encoder reranking.
& \texttt{sentence\_0}, \texttt{sentence\_1}, \texttt{label} \\
Generation
& \texttt{OTel-LLM}
& Generate grounded telecom answers from retrieved context.
& \texttt{prompt}, \texttt{completion}, \texttt{abstention}, chunk-count metadata \\
Safety
& \texttt{OTel-Safety}
& Abstain when appropriate.
& Schema-compatible with \texttt{OTel-LLM}; includes \texttt{abstention} and chunk-count metadata \\
\bottomrule
\end{tabularx}
\caption{Released OTel derived dataset family, ordered by the retrieval-reranking-generation workflow and its safety extension.}
\label{tab:dataset-family}
\end{table}

\begin{table}[!htbp]
\centering
\small
\setlength{\tabcolsep}{5pt}
\renewcommand{\arraystretch}{1.22}
\begin{tabularx}{\textwidth}{L{1.9cm} L{2.9cm} Y}
\toprule
RAG role & Dataset & Shortened real example \\
\midrule
Retrieval
& \texttt{OTel-Embedding}
& Retrieval example: a query about F1 Measurement ID Coordination is paired with the relevant F1-log passage and five nearby hard negatives. \\
Reranking
& \texttt{OTel-Reranker}
& Reranking example: an O-RAN Fronthaul Gateway query-passage pair about split 7-2 to 8 is marked relevant with \texttt{label}=1.0. \\
Generation
& \texttt{OTel-LLM}
& Generation example: an MBSFN-mode question is answered from retrieved context about \texttt{TDD MBSFN Information}. \\
Safety
& \texttt{OTel-Safety}
& Safety example: an SCP trust-domain question has off-topic retrieved contexts, so the completion abstains with \texttt{abstention}=true. \\
\bottomrule
\end{tabularx}
\caption{Compact examples drawn from released OTel rows and shortened for readability. Full rows and regeneration commands are documented in the release materials.}
\label{tab:dataset-examples}
\end{table}

\section{OTel Post-Trained Baselines}

The OTel model family is intended as a reproducible baseline suite for the community rather than a final word on telecom modeling. The release contains 30 full-parameter post-trained OTel baselines following the RAG workflow: 10 embedding models, 3 rerankers, and 17 language models. These baselines establish starting points for retrieval, reranking, and context-grounded generation; show the effect of OTel post-training across model sizes and architectures; and provide reference models for the community to beat, adapt, and extend.

Each released baseline starts from an open base checkpoint and is then full-parameter post-trained on the matching OTel dataset. The 10 embedding models use \texttt{OTel-Embedding}, the 3 rerankers use \texttt{OTel-Reranker}, and the 17 benchmarked LLMs use \texttt{OTel-LLM}. \texttt{OTel-Safety} is reserved for two abstention-focused auxiliary models, \texttt{OTel-LLM-8.3B-Safety} and \texttt{OTel-LLM-12B-Safety}, which are discussed in the appendix and are not counted in the 30-model baseline release.

We use the term \emph{full-parameter post-trained OTel baselines} deliberately: these are not frozen encoders, prompt-only wrappers, or loosely assembled checkpoints, but a controlled family of post-trained telecom models built with a shared recipe.

\begin{table}[t]
\centering
\small
\begin{tabular}{lcc}
\toprule
Family & Count & Primary dataset \\
\midrule
Embeddings & 10 & \texttt{OTel-Embedding} \\
Rerankers & 3 & \texttt{OTel-Reranker} \\
LLMs & 17 & \texttt{OTel-LLM} \\
\bottomrule
\end{tabular}
\caption{The 30 released full-parameter post-trained OTel baselines, ordered by the RAG workflow.}
\label{tab:family-counts}
\end{table}

A representative example is \texttt{OTel-LLM-8.3B-IT}. We initialize from the open \texttt{rnj-1-instruct} checkpoint~\citep{rnj1} and then full-parameter post-train on \texttt{OTel-LLM} for three epochs, producing a telecom-specialized instruction model. The same pattern is repeated across the released OTel baseline family: an open base model is selected, the family-appropriate OTel dataset is loaded, and the full parameter set is updated under a common configuration template.

The shared training recipe uses AdamW in 8-bit form, cosine decay with warmup, seed 42, a maximum sequence length of 1500 tokens, BF16 precision, Flash Attention 2, gradient checkpointing, and Fully Sharded Data Parallel training. LLMs and embedding models train for three epochs; rerankers train for two. Full base-model mappings and hyperparameter tables are deferred to the appendix to keep the main paper focused on the E\&D contribution rather than a file-by-file training inventory.

\section{Evaluation Protocol}

OTel evaluates three roles in the telecom RAG stack: embeddings for retrieval, rerankers for passage ordering, and LLMs for context-grounded answer generation. The evaluation protocol is designed to be reproducible and consistent with the training resources released alongside the baselines.

The LLM evaluation is scoped to context-grounded answer generation: models receive retrieved telecom context and are judged on whether their answers are correct relative to that context and the reference answer. These results should not be interpreted as unrestricted context-free QA performance; appendix-only GSMA leaderboard experiments cover that separate setting.

In this paper, safety refers to learning when retrieved context is insufficient or off-topic and the correct behavior is to abstain rather than answer.

\paragraph{Held-out evaluation splits.}
For LLMs and embeddings, the default split is 90\% train and 10\% eval with seed 42. For rerankers, the default split is 95\% train and 5\% eval with seed 42. Unless a stricter external benchmark split is added before submission, all main-paper results are reported on these held-out OTel evaluation partitions.

\paragraph{Metrics.}
LLMs are evaluated by LLM-as-judge correctness on context-grounded answer generation, using GPT-4o mini and Claude Sonnet 3.5 as judge models to determine whether a generated answer is correct given the provided context and reference answer; the reported correctness is the average score across the two judges. Embeddings are evaluated by NDCG@10, because retrieval quality depends on ranking the most relevant telecom passages at the top of the returned list. Rerankers are evaluated by MRR@10, because reranking quality is governed by how quickly the first truly relevant passage is promoted near the top. Where available, gains over the models without OTel fine-tuning are reported within the same model family.

\paragraph{Reporting policy.}
The main comparison is across the full-parameter post-trained OTel baselines within each family. All results include base-model performance before OTel fine-tuning and standard errors via bootstrap resampling ($n{=}1000$). Auxiliary experiments that are useful but not part of the core 30-model release, including the two abstention-focused safety variants, the TeleLogs classification head, the GSMA non-abstention QnA variant, and GPU utilization analysis, are reported in the appendix.

\paragraph{Collaborator-led stress tests.}
To probe behavior beyond the aggregate held-out splits, collaborators also ran auxiliary retrieval stress tests. UT Dallas and Purdue evaluated O-RAN retrieval on fixed-chunk question pools, while the University of Leeds examined retrieval behavior across 3GPP, GSMA PRD, industry whitepapers, O-RAN, RFCs, and telecom academic papers. These analyses are not reported as main quantitative benchmark tables because the protocols were not yet standardized across all domains, but they provide useful diagnostic evidence: O-RAN retrieval appears comparatively strong, whereas academic-paper and GSMA PRD examples remain weaker and need further curation.

\section{Results}

All results are reported on held-out OTel evaluation partitions. Standard errors are computed via bootstrap resampling (1,000 iterations). Each table includes performance without fine-tuning, performance with OTel fine-tuning, and the absolute gain from OTel fine-tuning.

\paragraph{Language-model baselines.}
Table~\ref{tab:llm-results} reports LLM-as-judge correctness for the 17 released instruction-tuned and reasoning LLM baselines. OTel fine-tuning yields consistent gains across all model sizes, ranging from +3.3 to +9.4 percentage points. \texttt{OTel-LLM-27B-IT} achieves the highest correctness at 87.8\%, while \texttt{OTel-LLM-8.3B-IT} is the strongest mid-size baseline at 79.0\%, outperforming other models in its weight class. Larger models generally raise the upper envelope, although architecture and training family still matter. \texttt{OTel-LLM-1.2B-IT} provides a competitive low-latency option at 73.8\%. Two abstention-focused safety variants are discussed separately in the appendix and are not counted in the initial 30-model release.

\begin{table}[!htbp]
\centering
\small
\begin{tabular}{llcccc}
\toprule
OTel model & Base model & Params (B) & \shortstack{Without\\fine-tuning} & \shortstack{With OTel\\fine-tuning} & $\Delta$ (pp) \\
\midrule
\texttt{OTel-LLM-270M-IT} & gemma-3-270m-it & 0.27 & 22.2 & \textbf{30.7} $\pm$ 1.4 & +8.5 \\
\texttt{OTel-LLM-0.6B-IT} & Qwen3-0.6B & 0.60 & 49.0 & \textbf{58.4} $\pm$ 1.1 & +9.4 \\
\texttt{OTel-LLM-1B-IT} & gemma-3-1b-it & 1.00 & 48.3 & \textbf{56.8} $\pm$ 1.0 & +8.5 \\
\texttt{OTel-LLM-1.2B-IT} & LFM2.5-1.2B-Instruct & 1.20 & 66.4 & \textbf{73.8} $\pm$ 0.8 & +7.4 \\
\texttt{OTel-LLM-1.7B-IT} & Qwen3-1.7B & 1.70 & 52.8 & \textbf{60.8} $\pm$ 0.9 & +8.0 \\
\texttt{OTel-LLM-3B-IT} & Mistral-3-3B & 3.00 & 56.9 & \textbf{63.9} $\pm$ 0.9 & +7.0 \\
\texttt{OTel-LLM-4B-IT} & gemma-3-4b-it & 4.00 & 66.2 & \textbf{72.7} $\pm$ 0.8 & +6.5 \\
\texttt{OTel-LLM-7B-IT} & OLMo-3-7B & 7.00 & 57.4 & \textbf{62.9} $\pm$ 0.9 & +5.5 \\
\texttt{OTel-LLM-8.2B-IT} & Qwen3-8B & 8.20 & 60.4 & \textbf{65.9} $\pm$ 0.8 & +5.5 \\
\texttt{OTel-LLM-8.3B-IT} & RNJ-1-Instruct & 8.30 & 72.5 & \textbf{79.0} $\pm$ 0.7 & +6.5 \\
\texttt{OTel-LLM-12B-IT} & gemma-3-12b-it & 12.00 & 78.3 & \textbf{82.8} $\pm$ 0.6 & +4.5 \\
\texttt{OTel-LLM-14B-IT} & Qwen3-14B & 14.00 & 60.7 & \textbf{65.7} $\pm$ 0.8 & +5.0 \\
\texttt{OTel-LLM-20B-IT} & GPT-OSS-20B & 20.00 & 61.4 & \textbf{65.9} $\pm$ 0.8 & +4.5 \\
\texttt{OTel-LLM-20B-Reasoning} & GPT-OSS-20B & 20.00 & 65.2 & \textbf{71.1} $\pm$ 0.8 & +5.9 \\
\texttt{OTel-LLM-24B-IT} & LFM2-24B-A2B & 24.00 & 75.0 & \textbf{79.0} $\pm$ 0.7 & +4.0 \\
\texttt{OTel-LLM-27B-IT} & gemma-3-27b-it & 27.00 & 84.5 & \textbf{87.8} $\pm$ 0.5 & +3.3 \\
\texttt{OTel-LLM-32B-IT} & OLMo-3-32B & 32.00 & 66.8 & \textbf{70.8} $\pm$ 0.8 & +4.0 \\
\bottomrule
\end{tabular}
\caption{LLM-as-judge correctness (\%) on held-out OTel evaluation splits. Correctness is the average score from GPT-4o mini and Claude Sonnet 3.5. OTel fine-tuning improves correctness across all language-model baselines, and larger models generally raise the upper envelope while architecture still matters. $\Delta$ is the absolute percentage-point gain. Standard errors via bootstrap ($n{=}1000$).}
\label{tab:llm-results}
\end{table}

\paragraph{Embedding baselines.}
Table~\ref{tab:embedding-results} reports NDCG@10 for the 10 embedding baselines. OTel fine-tuning on telecom-specific retrieval supervision produces large gains across all model sizes, with improvements ranging from +9.2 to +59.7 percentage points over the base models. Even the smallest model (0.022B parameters) reaches 83.8\% NDCG@10 after OTel fine-tuning, while the largest (8.000B) achieves 93.1\%.

\begin{table}[!htbp]
\centering
\small
\begin{tabular}{llcccc}
\toprule
OTel model & Base model & Params (B) & \shortstack{Without\\fine-tuning} & \shortstack{With OTel\\fine-tuning} & $\Delta$ (pp) \\
\midrule
\texttt{OTel-Embedding-22M} & all-MiniLM-L6-v2 & 0.022 & 24.1 & \textbf{83.8} $\pm$ 0.8 & +59.7 \\
\texttt{OTel-Embedding-33M} & bge-small-en-v1.5 & 0.033 & 31.4 & \textbf{86.3} $\pm$ 0.7 & +54.9 \\
\texttt{OTel-Embedding-34M} & all-MiniLM-L12-v2 & 0.034 & 29.2 & \textbf{84.6} $\pm$ 0.8 & +55.4 \\
\texttt{OTel-Embedding-109M} & all-mpnet-base-v2 & 0.109 & 38.5 & \textbf{87.2} $\pm$ 0.7 & +48.7 \\
\texttt{OTel-Embedding-300M} & Gemma3-Embedding-300M & 0.300 & 72.3 & \textbf{90.4} $\pm$ 0.6 & +18.1 \\
\texttt{OTel-Embedding-335M} & bge-large-en-v1.5 & 0.335 & 51.7 & \textbf{89.2} $\pm$ 0.6 & +37.5 \\
\texttt{OTel-Embedding-568M} & bge-m3 & 0.568 & 57.2 & \textbf{89.6} $\pm$ 0.6 & +32.4 \\
\texttt{OTel-Embedding-600M} & Qwen3-Embedding-0.6B & 0.600 & 79.7 & \textbf{90.0} $\pm$ 0.5 & +10.3 \\
\texttt{OTel-Embedding-4B} & Qwen3-Embedding-4B & 4.000 & 82.5 & \textbf{91.7} $\pm$ 0.5 & +9.2 \\
\texttt{OTel-Embedding-8B} & Qwen3-Embedding-8B & 8.000 & 83.9 & \textbf{93.1} $\pm$ 0.4 & +9.2 \\
\bottomrule
\end{tabular}
\caption{NDCG@10 (\%) on held-out OTel evaluation splits. OTel fine-tuning improves telecom retrieval quality across embedding sizes. $\Delta$ is the absolute NDCG@10 percentage-point gain. Standard errors via bootstrap ($n{=}1000$).}
\label{tab:embedding-results}
\end{table}

\paragraph{Reranker baselines.}
Table~\ref{tab:reranker-results} reports MRR@10 for the three reranker baselines. All three models achieve MRR@10 between 0.938 and 0.947 after OTel fine-tuning, up from 0.346--0.417 for the base models, a gain of 0.530--0.592 in absolute MRR@10.

\begin{table}[t]
\centering
\small
\begin{tabular}{lcccc}
\toprule
OTel model & Params (B) & \shortstack{Without\\fine-tuning} & \shortstack{With OTel\\fine-tuning} & $\Delta$ \\
\midrule
\texttt{OTel-Reranker-0.6B} & 0.60 & 0.346 & \textbf{0.938} $\pm$ 0.007 & +0.592 \\
\texttt{OTel-Reranker-4B} & 4.00 & 0.407 & \textbf{0.943} $\pm$ 0.006 & +0.536 \\
\texttt{OTel-Reranker-8B} & 8.00 & 0.417 & \textbf{0.947} $\pm$ 0.005 & +0.530 \\
\bottomrule
\end{tabular}
\caption{MRR@10 on held-out OTel evaluation splits. OTel fine-tuning improves reranking quality for all three reranker baselines. $\Delta$ is the absolute MRR@10 gain. Standard errors via bootstrap ($n{=}1000$).}
\label{tab:reranker-results}
\end{table}

\paragraph{End-to-end interpretation.}
Taken together, the retrieval, reranking, and generation baselines allow practitioners to choose a telecom RAG stack based on the deployment constraint that matters most. The base-model comparisons in Tables~\ref{tab:llm-results}--\ref{tab:reranker-results} confirm that OTel fine-tuning consistently improves domain performance across all three model families and all parameter scales. Larger models generally improve the upper envelope, while architecture and training family still influence the final ranking. Smaller models remain useful for latency-sensitive deployments, and mid-size models provide strong cost-performance trade-offs.

\begin{figure}[!htbp]
\centering
\IfFileExists{plots/LLM_performance.jpeg}{\includegraphics[width=0.56\linewidth]{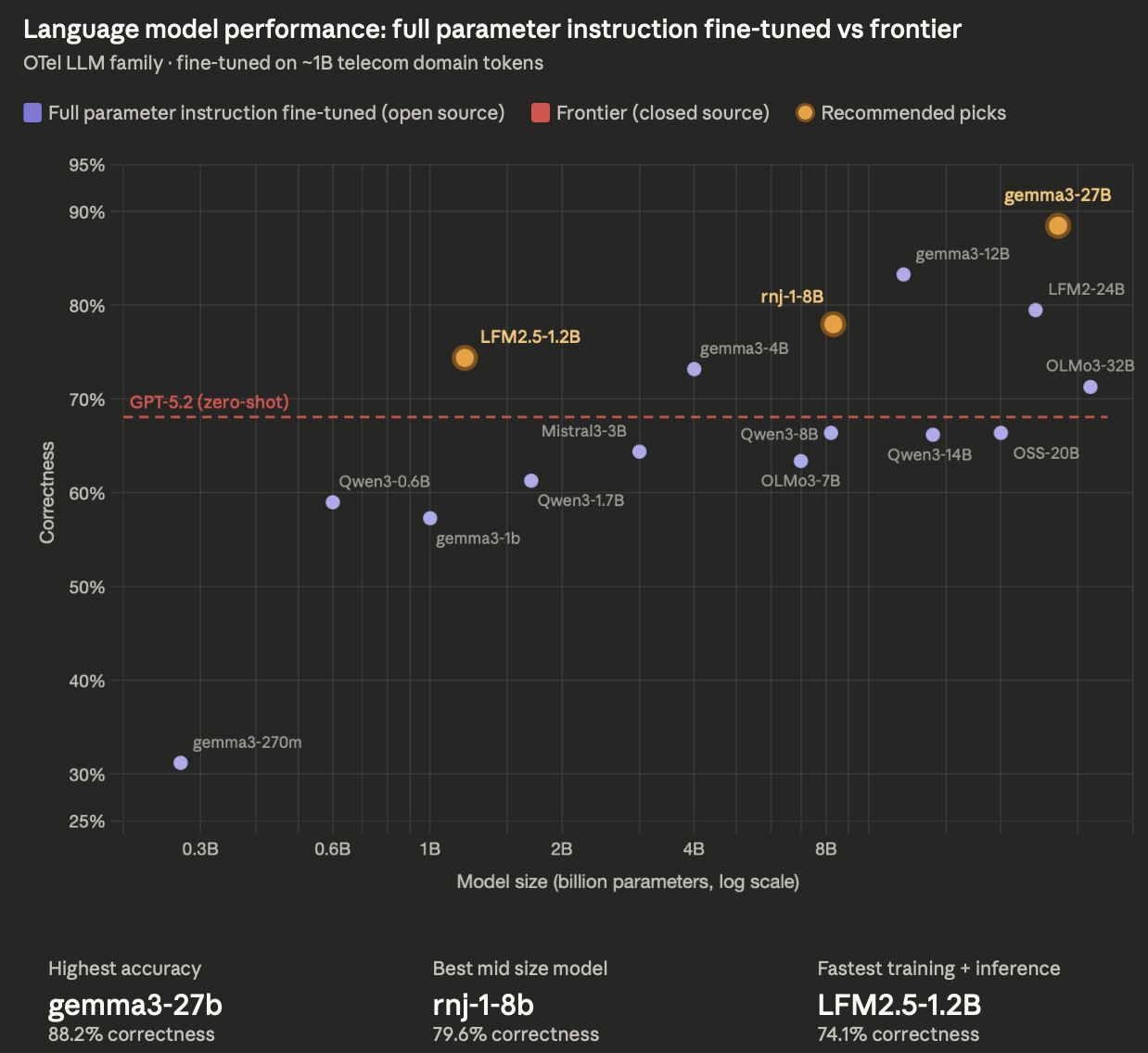}}{\fbox{\parbox{0.75\linewidth}{\centering Placeholder for plots/LLM\_performance.jpeg}}}

\vspace{1pt}

\IfFileExists{plots/Embedding_performance.png}{\includegraphics[width=0.56\linewidth]{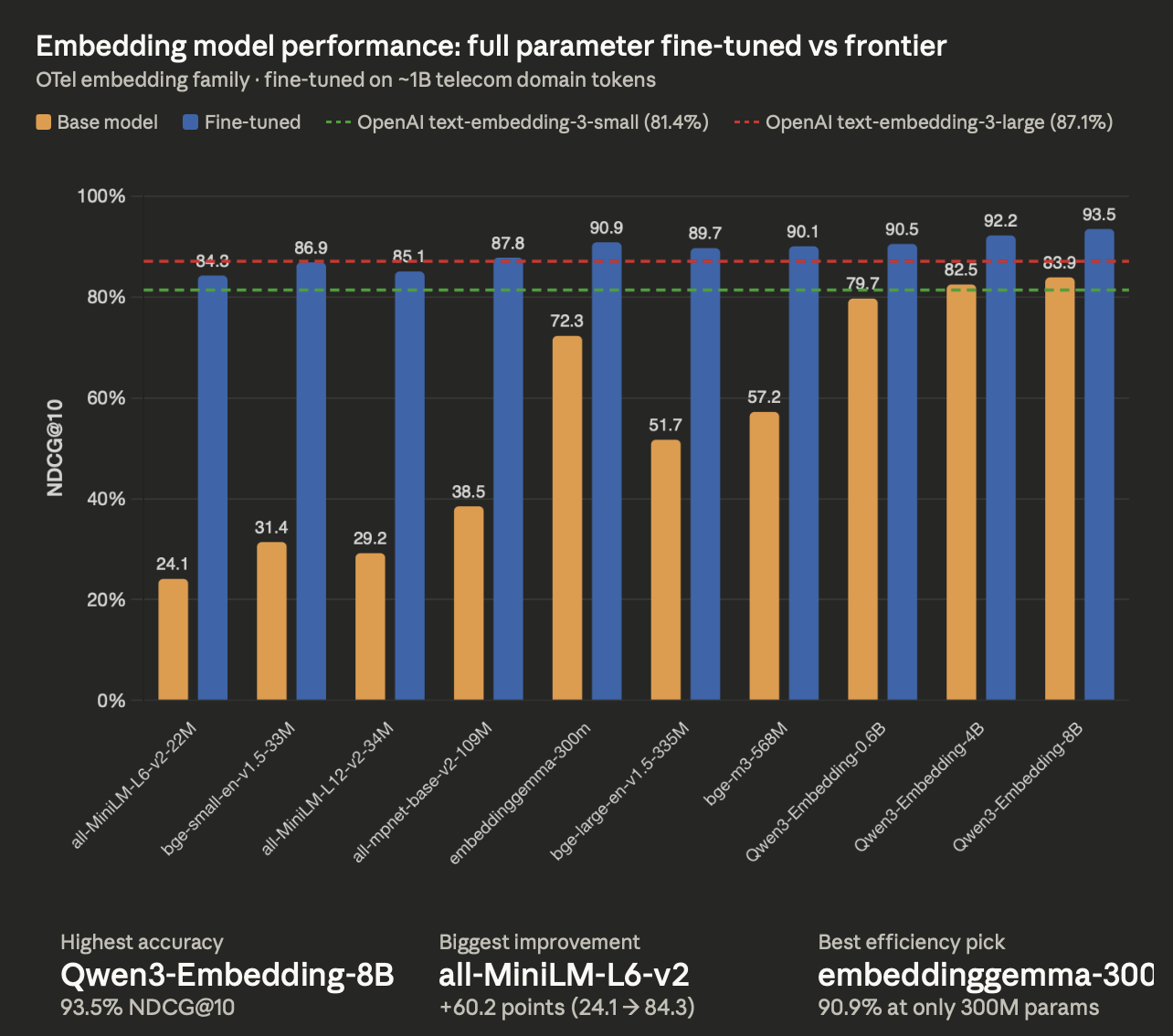}}{\fbox{\parbox{0.75\linewidth}{\centering Placeholder for plots/Embedding\_performance.png}}}
\caption{OTel fine-tuning improves language-model correctness and embedding retrieval quality; larger LLMs generally raise the correctness upper envelope.}
\label{fig:results-summary}
\end{figure}

\section{Limitations, Broader Impact, and Responsible Release}

OTel is intentionally domain-specific. The reported results should not be generalized outside telecommunications, and even within telecom the current resource remains English-only and primarily text-centric. The OTel LLMs are designed for context-grounded RAG, which means they are not optimized for unrestricted open-ended QA. The main-paper evaluation also relies on held-out splits drawn from the released OTel resource family rather than a fully independent external benchmark suite, so future work should expand the amount of external testing and cross-benchmark transfer analysis.

The collaborator-led retrieval stress tests also indicate that aggregate scores hide important subdomain variation. O-RAN retrieval appears comparatively strong, while academic-paper and GSMA PRD examples remain weaker areas. Future releases should expand and re-clean the academic-paper subset, re-curate GSMA PRD examples with harder negatives and larger evaluation pools, and make per-subdomain reporting a first-class part of the OTel evaluation protocol.

Source coverage is broad but still imperfect. Some telecom subdomains are better represented than others, and the source material mixes standards, reference documents, whitepapers, academic papers, and web-derived data. That heterogeneity is valuable for realism, but it creates nontrivial provenance and redistribution work. For this reason, release documentation should describe source classes, intended use, and source-specific constraints rather than oversimplify the upstream licensing story.

The resource has clear positive value: it reduces barriers to open telecom experimentation, enables reproducible comparison across retrievers, rerankers, and generators, and offers a public starting point for telecom-specific model development. The main risks are misuse of domain-specialized models to generate misleading telecom content and over-trust in models outside their intended RAG setting. Therefore, abstention-oriented safety tuning, model cards, dataset cards, intended-use statements, and asset-level metadata are critical for responsible release.

\section{Conclusion}

OTel delivers the resource promised at the start of the paper: documented telecom data sources and tasks, a derived dataset family, held-out evaluation partitions, and Hugging Face reference baselines spanning retrieval, reranking, and context-grounded generation. The resource has already seen substantial community use, with more than 16 million model downloads and 157+ pieces of media coverage worldwide, and the released baselines provide reference points for the community to beat, adapt, and extend.

The results show strong headline performance: 93.1\% NDCG@10 for retrieval, 0.947 MRR@10 for reranking, and 87.8\% LLM correctness. OTel fine-tuning improves performance across embedding models, rerankers, and LLMs, and model scale helps raise the upper envelope while architecture still matters. The goal is not to claim that telecom is ``solved,'' but to provide an open and reproducible starting point for the community to expand the data, improve retrieval and reranking, build stronger context-grounded telecom LLMs, and extend OTel through multilingual, multimodal, per-subdomain, and external benchmark evaluations.

\section*{Acknowledgment}

We thank Mashroor Hasan Bhuiyan (The University of Texas at Dallas); Kostikey Mustakas (AT\&T); Nikolaos Vasiloglou and Molham Aref (RelationalAI); Enrique Molero (GSMA); Merouane Abdelkader Debbah, Esraa Fahmy, and Bohao Wang (Khalifa University); Zeinab Nezami, Maryam Hafeez, and Shehr Bano (University of Leeds); Leandros Tassiulas and Rex Ying (Yale University); Somanshu Singla and Adarsh Chaluvaraju (Essential AI); Alexander Finn, Andy Allred, Andrey Ivannikov, Antti-Ville Suni, Mark van Heeswijk, and Kumaran Siva (AMD); Matt Upson and Nick Sorros (Mantis NLP); Faik Kerem Ors and Mirza Masfiqur Rahman (Purdue University); and Vignesh Ethiraj and Ashwath David (NetoAI) for their contributions to the OTel project. 

\small
\bibliographystyle{plainnat}
\bibliography{otel}

\appendix

\section{Full Baseline Roster}

\subsection{Language models}

\begin{table}[H]
\centering
\small
\resizebox{\textwidth}{!}{
\begin{tabular}{lll}
\toprule
OTel model & Parameters & Base model \\
\midrule
\texttt{OTel-LLM-270M-IT} & 270M & \texttt{gemma-3-270m-it} \\
\texttt{OTel-LLM-0.6B-IT} & 0.6B & \texttt{Qwen3-0.6B} \\
\texttt{OTel-LLM-1B-IT} & 1B & \texttt{gemma-3-1b-it} \\
\texttt{OTel-LLM-1.2B-IT} & 1.2B & \texttt{LFM2.5-1.2B-Instruct} \\
\texttt{OTel-LLM-1.7B-IT} & 1.7B & \texttt{Qwen3-1.7B} \\
\texttt{OTel-LLM-3B-IT} & 3B & \texttt{Mistral-3-3B} \\
\texttt{OTel-LLM-4B-IT} & 4B & \texttt{gemma-3-4b-it} \\
\texttt{OTel-LLM-7B-IT} & 7B & \texttt{OLMo-3-7B} \\
\texttt{OTel-LLM-8.2B-IT} & 8.2B & \texttt{Qwen3-8B} \\
\texttt{OTel-LLM-8.3B-IT} & 8.3B & \texttt{RNJ-1-Instruct} \\
\texttt{OTel-LLM-12B-IT} & 12B & \texttt{gemma-3-12b-it} \\
\texttt{OTel-LLM-14B-IT} & 14B & \texttt{Qwen3-14B} \\
\texttt{OTel-LLM-20B-IT} & 20B & \texttt{GPT-OSS-20B} \\
\texttt{OTel-LLM-20B-Reasoning} & 20B & \texttt{GPT-OSS-20B} \\
\texttt{OTel-LLM-24B-IT} & 24B & \texttt{LFM2-24B-A2B} \\
\texttt{OTel-LLM-27B-IT} & 27B & \texttt{gemma-3-27b-it} \\
\texttt{OTel-LLM-32B-IT} & 32B & \texttt{OLMo-3-32B} \\
\bottomrule
\end{tabular}
}
\caption{Language-model subset of the 30 released full-parameter post-trained OTel baselines.}
\end{table}

\subsection{Embedding models}

\begin{table}[H]
\centering
\small
\resizebox{\textwidth}{!}{
\begin{tabular}{lll}
\toprule
OTel model & Parameters & Base model \\
\midrule
\texttt{OTel-Embedding-22M} & 22M & \texttt{all-MiniLM-L6-v2} \\
\texttt{OTel-Embedding-33M} & 33M & \texttt{BAAI/bge-small-en-v1.5} \\
\texttt{OTel-Embedding-34M} & 34M & \texttt{all-MiniLM-L12-v2} \\
\texttt{OTel-Embedding-109M} & 109M & \texttt{all-mpnet-base-v2} \\
\texttt{OTel-Embedding-300M} & 300M & \texttt{Gemma3-Embedding-300M} \\
\texttt{OTel-Embedding-335M} & 335M & \texttt{BAAI/bge-large-en-v1.5} \\
\texttt{OTel-Embedding-568M} & 568M & \texttt{BAAI/bge-m3} \\
\texttt{OTel-Embedding-600M} & 600M & \texttt{Qwen3-Embedding-0.6B} \\
\texttt{OTel-Embedding-4B} & 4B & \texttt{Qwen3-Embedding-4B} \\
\texttt{OTel-Embedding-8B} & 8B & \texttt{Qwen3-Embedding-8B} \\
\bottomrule
\end{tabular}
}
\caption{Embedding subset of the 30 released full-parameter post-trained OTel baselines.}
\end{table}

\subsection{Reranker models}

\begin{table}[H]
\centering
\small
\begin{tabular}{lll}
\toprule
OTel model & Parameters & Base model \\
\midrule
\texttt{OTel-Reranker-0.6B} & 0.6B & \texttt{Qwen3-0.6B} \\
\texttt{OTel-Reranker-4B} & 4B & \texttt{Qwen3-4B} \\
\texttt{OTel-Reranker-8B} & 8B & \texttt{Qwen3-8B} \\
\bottomrule
\end{tabular}
\caption{Reranker subset of the 30 released full-parameter post-trained OTel baselines.}
\end{table}

\section{Training Datasets}

The four released OTel datasets support the 30-model initial release plus appendix-only auxiliary safety analyses. In particular, \texttt{OTel-Safety} is used for the two abstention-focused variants discussed in Appendix~H.

\begin{table}[H]
\centering
\small
\begin{tabular}{ll}
\toprule
Model type & Dataset \\
\midrule
LLM instruction tuning & \texttt{OTel-LLM} \\
LLM safety tuning & \texttt{OTel-Safety} \\
Embedding & \texttt{OTel-Embedding} \\
Reranker & \texttt{OTel-Reranker} \\
\bottomrule
\end{tabular}
\caption{Dataset mapping for the OTel training releases.}
\end{table}

\section{Training Hyperparameters and Optimization}

\begin{table}[H]
\centering
\small
\begin{tabular}{ll}
\toprule
Parameter & Value \\
\midrule
Optimizer & AdamW (8-bit, bitsandbytes) \\
Learning-rate schedule & Cosine decay with warmup \\
Weight decay & 0.01 \\
Warmup steps & 100 \\
Per-device batch size & 8--128 \\
Gradient accumulation & 4--32 \\
Random seed & 42 \\
Maximum sequence length & 1500 \\
Training epochs (LLM, embedding) & 3 \\
Training epochs (reranker) & 2 \\
\bottomrule
\end{tabular}
\caption{Shared optimization template for the full-parameter post-trained OTel baselines.}
\end{table}

\section{Memory, Precision, and Distributed Training}

\begin{table}[H]
\centering
\small
\begin{tabular}{ll}
\toprule
Parameter & Value \\
\midrule
Numerical precision & BF16 \\
Attention implementation & Flash Attention 2 \\
Gradient checkpointing & Enabled \\
Distributed training & Fully Sharded Data Parallel \\
\bottomrule
\end{tabular}
\caption{Precision and distributed training configuration.}
\end{table}

\section{Compute Resources}

\begin{table}[H]
\centering
\small
\begin{tabular}{lr}
\toprule
Hardware & Count \\
\midrule
AMD MI300X & 32 \\
AMD MI325X & 64 \\
AMD MI355X & 32 \\
NVIDIA A100 & 8 \\
NVIDIA H100 & 32 \\
\bottomrule
\end{tabular}
\caption{GPU pools used across OTel training runs.}
\end{table}

\section{Dataset Splits and Logging}

\begin{table}[H]
\centering
\small
\begin{tabular}{llll}
\toprule
Model type & Train & Eval & Seed \\
\midrule
LLM & 90\% & 10\% & 42 \\
Embedding & 90\% & 10\% & 42 \\
Reranker & 95\% & 5\% & 42 \\
\bottomrule
\end{tabular}
\caption{Held-out split policy used in the current OTel evaluation setup.}
\end{table}

\begin{table}[H]
\centering
\small
\begin{tabular}{ll}
\toprule
Logging/configuration item & Value \\
\midrule
Evaluation strategy & Per epoch \\
Save strategy & Per epoch \\
Logging interval & Every 50 steps \\
Experiment tracking & TensorBoard \\
LLM judge models & GPT-4o mini and Claude Sonnet 3.5 \\
\bottomrule
\end{tabular}
\caption{Evaluation cadence and experiment logging defaults.}
\end{table}

\section{Loss Computation}

\begin{table}[H]
\centering
\small
\begin{tabular}{ll}
\toprule
Parameter & Value \\
\midrule
Loss masking & Completion tokens only \\
Label smoothing & None \\
Padding side & Right \\
\bottomrule
\end{tabular}
\caption{Shared loss-side configuration.}
\end{table}

\section{Additional Experiments Outside the Core 30-Model Release}

The following experiments are useful auxiliary validations but are not part of the core 30-model OTel release described in the main paper.

\paragraph{Appendix-only safety variants.}
Two full-parameter post-trained abstention-focused variants, \texttt{OTel-LLM-8.3B-Safety} and \texttt{OTel-LLM-12B-Safety}, were trained on \texttt{OTel-Safety}. They were not included in the initial 30-model release, so they are discussed here as auxiliary models rather than counted baseline results.

\paragraph{TeleLogs classification.}
A classification head added to the \texttt{rnj-1} base path reaches roughly 99\% accuracy on the TeleLogs 5G root-cause-analysis benchmark \citep{telelogs}. The resulting classification model weights and training code are publicly released through the OTel Hugging Face and GitHub repositories.

\paragraph{GSMA QnA comparison.}
One non-abstention QnA variant, \texttt{OTel-LLM-8.3B-QnA}, was trained after the initial release for direct comparison on context-free telecom QA benchmarks. This model starts from the same base checkpoint (\texttt{RNJ-1-Instruct}) as \texttt{OTel-LLM-8.3B-IT}, but follows a separate training path: continued pretraining on telecom documents, followed by RL and supervised fine-tuning on a broad telecom QnA mixture that includes, but is not limited to, \texttt{OTel-LLM}. This auxiliary variant is separate from \texttt{OTel-Safety} and is designed to answer direct questions without requiring RAG context, making it compatible with the GSMA Open Telco AI Leaderboard evaluation protocol. It is intentionally separated from the main paper because it is not part of the core RAG-oriented OTel baseline family.

\begin{figure}[!htbp]
\centering
\IfFileExists{plots/GSMA_leaderboard.png}{\includegraphics[width=0.95\linewidth,height=0.30\textheight,keepaspectratio]{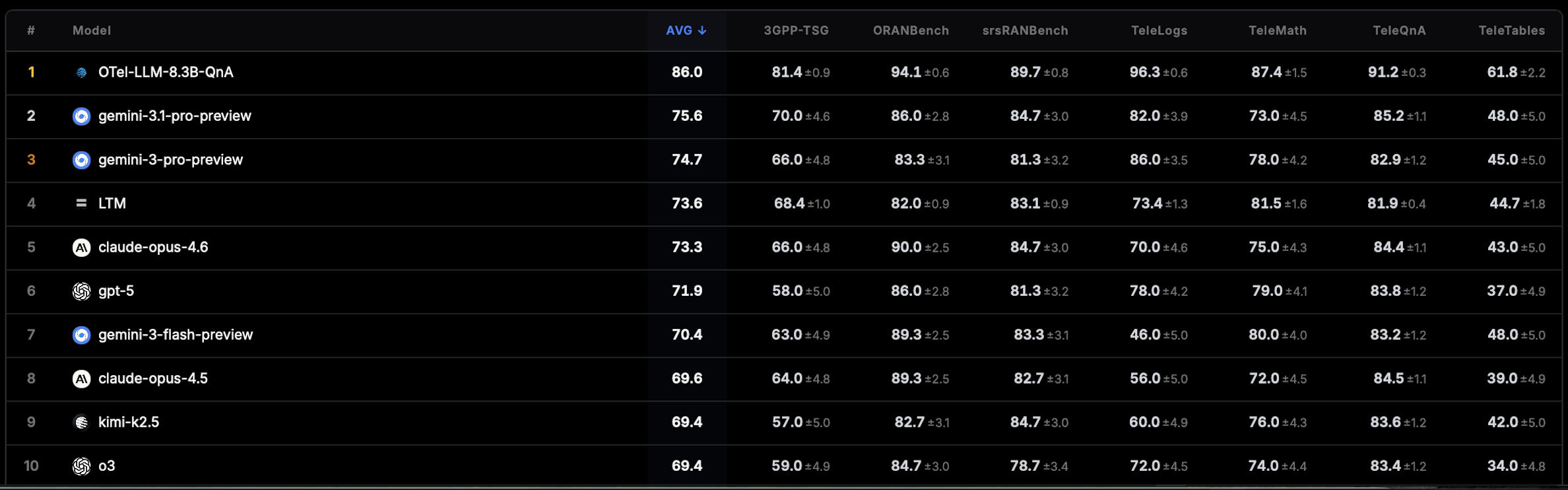}}{\fbox{\parbox{0.75\linewidth}{\centering Placeholder for plots/GSMA\_leaderboard.png}}}
\caption{Auxiliary GSMA Open Telco AI Leaderboard comparison for the non-abstention \texttt{OTel-LLM-8.3B-QnA} variant.}
\label{fig:gsma-leaderboard}
\end{figure}

\paragraph{GPU utilization.}
Large-scale training runs reached 94.2\% utilization across 256 AMD MI325X GPUs, as summarized in the accompanying utilization plot. We treat this as infrastructure evidence rather than a central E\&D claim.

\begin{figure}[!htbp]
\centering
\IfFileExists{plots/GPU_utilization.jpeg}{\includegraphics[width=0.88\linewidth,height=0.46\textheight,keepaspectratio]{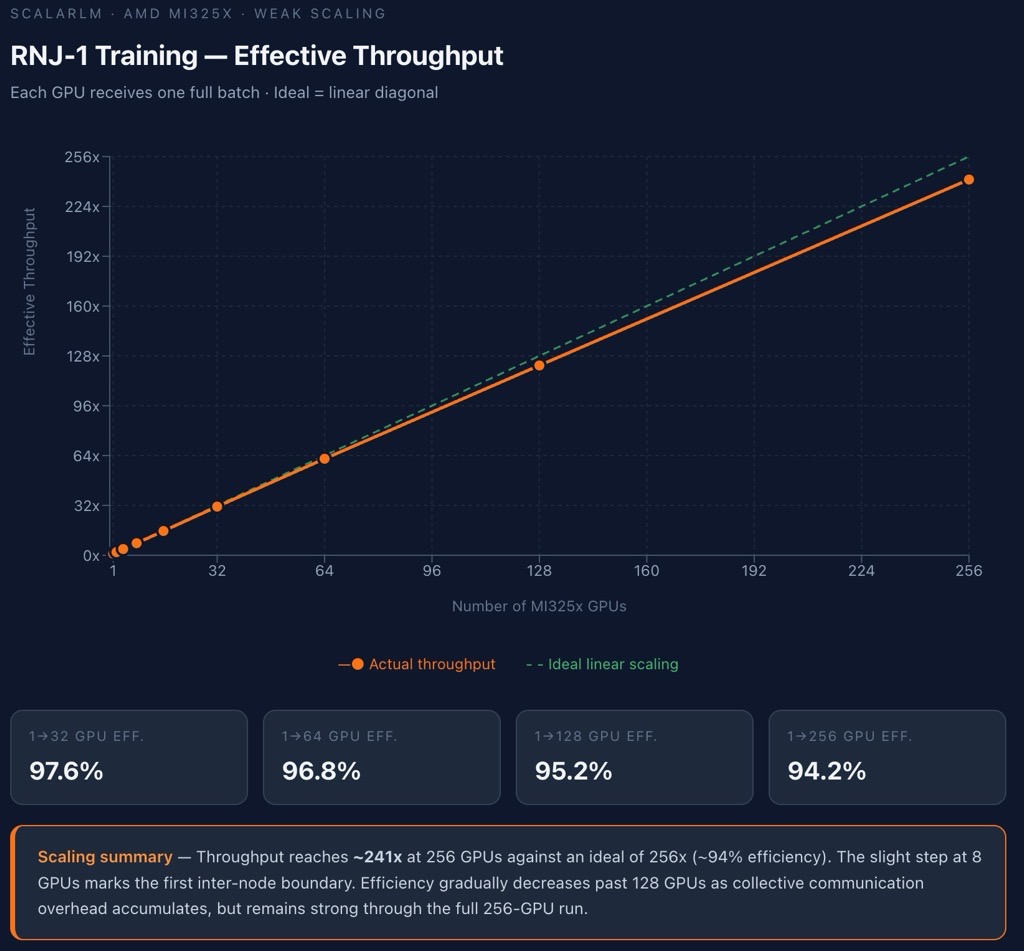}}{\fbox{\parbox{0.75\linewidth}{\centering Placeholder for plots/GPU\_utilization.jpeg}}}
\caption{Auxiliary GPU utilization analysis kept in the appendix rather than the main paper.}
\end{figure}

\section{Asset Documentation and Release Metadata}

The release package includes dataset cards, model cards, Croissant metadata files (with Responsible AI extensions), and reproduction instructions for the four datasets and 30 released full-parameter post-trained OTel baselines. All assets are publicly available on Hugging Face and GitHub. Each dataset repository includes a \texttt{croissant.json} file with validated Croissant 1.1 metadata and all required RAI fields (\texttt{rai:dataLimitations}, \texttt{rai:dataBiases}, \texttt{rai:personalSensitiveInformation}, \texttt{rai:dataUseCases}, \texttt{rai:dataSocialImpact}, \texttt{rai:hasSyntheticData}, and \texttt{prov:wasGeneratedBy}).

Public dataset and model cards should cite this paper as published at NeurIPS 2026 (Evaluations \& Datasets Track).

\begin{table}[H]
\centering
\small
\begin{tabular}{llll}
\toprule
Asset class & Release unit & Documentation & Metadata \\
\midrule
Dataset & \texttt{OTel-LLM} & Dataset card & Croissant + RAI \\
Dataset & \texttt{OTel-Safety} & Dataset card & Croissant + RAI \\
Dataset & \texttt{OTel-Embedding} & Dataset card & Croissant + RAI \\
Dataset & \texttt{OTel-Reranker} & Dataset card & Croissant + RAI \\
Model family & 30 released OTel baselines & Model cards & Reproduction instructions \\
Code & Training/evaluation scripts & README + commands & Environment specification \\
\bottomrule
\end{tabular}
\caption{Release artifacts and their accompanying documentation.}
\end{table}

\paragraph{Source provenance and licensing.}
All source documents used in the OTel dataset are publicly available and free to download. The OTel datasets release only derived QA pairs, not the raw source documents, under the Apache-2.0 license.

\begin{table}[H]
\centering
\small
\begin{tabularx}{\textwidth}{L{3.6cm} Y L{3.2cm}}
\toprule
Source class & Sources & Availability \\
\midrule
Standards body & 3GPP specifications & Public \\
Consortium documents & GSMA PRDs, Discover portal & Public \\
O-RAN documentation & O-RAN Alliance specifications & Public \\
Open technical corpora & IETF RFC series & Public \\
Academic papers & arXiv telecom papers & Public (CC-BY) \\
Web-derived & Common Crawl telecom pages & Public \\
Encyclopedia & Wikipedia telecom articles & Public (CC-BY-SA) \\
Industry whitepapers & Telecom vendor whitepapers & Public \\
\bottomrule
\end{tabularx}
\caption{Source provenance. OTel releases only derived QA pairs, under the Apache-2.0 license.}
\end{table}

\section{Worldwide Media Coverage of the Open Telco AI Project}

The project has received worldwide media coverage across telecom trade press, general technology outlets, regional business press, and syndicated pickups. The maintained source-of-truth list is provided in the supplemental appendix file \texttt{OTel\_Media\_Coverage\_List.md}, which enumerates the currently tracked outlets, regions, dates, headlines, and links supporting the media-coverage count cited in the main paper.

\IfFileExists{otel_checklist.tex}{\section*{NeurIPS Paper Checklist}

\begin{enumerate}

\item {\bf Claims}
    \item[] Question: Do the main claims made in the abstract and introduction accurately reflect the paper's contributions and scope?
    \item[] Answer: \answerYes{}
    \item[] Justification: The abstract and introduction state that the contribution is a telecom evaluation resource consisting of curated datasets, 30 released full-parameter post-trained OTel baselines, and a reproducible evaluation setup; see Sections~1--2.

\item {\bf Limitations}
    \item[] Question: Does the paper discuss the limitations of the work performed by the authors?
    \item[] Answer: \answerYes{}
    \item[] Justification: Section~7 discusses telecom-only scope, English-only coverage, text-centric evaluation, RAG-specific design choices, and licensing and redistribution constraints.

\item {\bf Theory assumptions and proofs}
    \item[] Question: For each theoretical result, does the paper provide the full set of assumptions and a complete (and correct) proof?
    \item[] Answer: \answerNA{}
    \item[] Justification: The paper does not present new theoretical results or proofs.

\item {\bf Experimental result reproducibility}
    \item[] Question: Does the paper fully disclose all the information needed to reproduce the main experimental results of the paper to the extent that it affects the main claims and/or conclusions of the paper (regardless of whether the code and data are provided or not)?
    \item[] Answer: \answerYes{}
    \item[] Justification: Sections~3--6 and Appendices~A--I describe dataset construction, train/eval splits, model families, optimization settings, compute resources, and release artifacts needed to reproduce the reported experiments.

\item {\bf Open access to data and code}
    \item[] Question: Does the paper provide open access to the data and code, with sufficient instructions to faithfully reproduce the main experimental results, as described in supplemental material?
    \item[] Answer: \answerYes{}
    \item[] Justification: The paper describes release of OTel datasets, final model weights, evaluation code, reproduction scripts, and accompanying cards/metadata; see Appendix~I.

\item {\bf Experimental setting/details}
    \item[] Question: Does the paper specify all the training and test details (e.g., data splits, hyperparameters, how they were chosen, type of optimizer) necessary to understand the results?
    \item[] Answer: \answerYes{}
    \item[] Justification: Section~5 gives the held-out split policy and metrics, while Appendices~C--G provide optimization, precision, distributed training, compute, loss masking, and logging details.

\item {\bf Experiment statistical significance}
    \item[] Question: Does the paper report error bars suitably and correctly defined or other appropriate information about the statistical significance of the experiments?
    \item[] Answer: \answerYes{}
    \item[] Justification: All result tables (Tables 3--5) report standard errors computed via bootstrap resampling ($n{=}10$). The standard errors capture variability across bootstrap samples of the held-out evaluation partitions.

\item {\bf Experiments compute resources}
    \item[] Question: For each experiment, does the paper provide sufficient information on the computer resources (type of compute workers, memory, time of execution) needed to reproduce the experiments?
    \item[] Answer: \answerYes{}
    \item[] Justification: Appendix~E summarizes the GPU pools used across AMD and NVIDIA hardware, and Sections~4--5 explain the common training configuration and evaluation cadence.

\item {\bf Code of ethics}
    \item[] Question: Does the research conducted in the paper conform, in every respect, with the NeurIPS Code of Ethics \url{https://neurips.cc/public/EthicsGuidelines}?
    \item[] Answer: \answerYes{}
    \item[] Justification: The work uses public technical sources, documents limitations and misuse risks, and discusses responsible release considerations in Section~7 and Appendix~I.

\item {\bf Broader impacts}
    \item[] Question: Does the paper discuss both potential positive societal impacts and negative societal impacts of the work performed?
    \item[] Answer: \answerYes{}
    \item[] Justification: Section~7 discusses both the potential benefits of open telecom evaluation resources and the risks of misuse or misleading generated telecom content.

\item {\bf Safeguards}
    \item[] Question: Does the paper describe safeguards that have been put in place for responsible release of data or models that have a high risk for misuse (e.g., pre-trained language models, image generators, or scraped datasets)?
    \item[] Answer: \answerYes{}
    \item[] Justification: The paper describes abstention-focused safety training, intended-use restrictions, and planned asset documentation and release metadata in Sections~4 and~7 and Appendix~I.

\item {\bf Licenses for existing assets}
    \item[] Question: Are the creators or original owners of assets (e.g., code, data, models), used in the paper, properly credited and are the license and terms of use explicitly mentioned and properly respected?
    \item[] Answer: \answerYes{}
    \item[] Justification: Appendix~I provides a source-by-source provenance and licensing table. All source documents are publicly available. The OTel datasets release only derived QA pairs under Apache-2.0, not raw source documents. Base models are credited with citations and their licenses are respected.

\item {\bf New assets}
    \item[] Question: Are new assets introduced in the paper well documented and is the documentation provided alongside the assets?
    \item[] Answer: \answerYes{}
    \item[] Justification: The paper introduces four dataset releases and 30 released full-parameter post-trained OTel baselines. Each dataset includes a dataset card, a validated Croissant 1.1 metadata file with all required Responsible AI fields, and reproduction documentation; see Appendix~I.

\item {\bf Crowdsourcing and research with human subjects}
    \item[] Question: For crowdsourcing experiments and research with human subjects, does the paper include the full text of instructions given to participants and screenshots, if applicable, as well as details about compensation (if any)?
    \item[] Answer: \answerNA{}
    \item[] Justification: The paper does not involve crowdsourcing experiments or research with human subjects.

\item {\bf Institutional review board (IRB) approvals or equivalent for research with human subjects}
    \item[] Question: Does the paper describe potential risks incurred by study participants, whether such risks were disclosed to the subjects, and whether Institutional Review Board (IRB) approvals (or an equivalent approval/review based on the requirements of your country or institution) were obtained?
    \item[] Answer: \answerNA{}
    \item[] Justification: The paper does not involve human subjects research.

\item {\bf Declaration of LLM usage}
    \item[] Question: Does the paper describe the usage of LLMs if it is an important, original, or non-standard component of the core methods in this research? Note that if the LLM is used only for writing, editing, or formatting purposes and does \emph{not} impact the core methodology, scientific rigor, or originality of the research, declaration is not required.
    \item[] Answer: \answerYes{}
    \item[] Justification: Section~3 describes LLM-assisted answer recreation and fact-grounded verification during data preparation, which is a non-standard methodological component of the resource construction pipeline.

\end{enumerate}
}{\section*{NeurIPS Checklist}\textit{Checklist file not included in this workspace. Add otel\_checklist.tex before final submission if required.}}

\end{document}